\documentclass{article}

\usepackage[final]{acl}

\usepackage{times}
\usepackage{latexsym}
\usepackage[T1]{fontenc}
\usepackage[utf8]{inputenc}
\usepackage{microtype}
\usepackage{inconsolata}
\usepackage{graphicx}
\usepackage{dsfont}
\usepackage{amsmath}

\usepackage{amsmath,amssymb,amsfonts,amsthm}
\usepackage{bbm}
\usepackage{booktabs}
\usepackage{hyperref}
\usepackage{url}
\usepackage[capitalize]{cleveref}
\usepackage{algorithm}
\usepackage{algorithmic}
\usepackage{xcolor}
\usepackage{verbatim}
\usepackage[textsize=scriptsize]{todonotes}

\usepackage{caption}
\usepackage{booktabs}
\usepackage{textcomp}
\usepackage{tcolorbox}
\newcommand{\q}{x} %
\newcommand{\B}[1]{\textbf{#1}} %
\newcommand{\stdv}[1]{\textsubscript{\scriptsize $\pm$#1}} %

\newtheorem{theorem}{Theorem}

\newcommand{\V}{\textcolor[RGB]{34,139,34}{\boldsymbol{p_V}(v {=} 1 \mid x, y)}}
\newcommand{\Vother}{\textcolor[RGB]{34,139,34}{\boldsymbol{p_V}(y \mid x)}}
\newcommand{\Vbare}{\textcolor[RGB]{34,139,34}{\boldsymbol{p_V}}}
\newcommand{\G}{\textcolor[RGB]{30,100,200}{\boldsymbol{p_G}(y \mid x)}}
\newcommand{\Gprime}{\textcolor[RGB]{30,100,200}{\boldsymbol{p_{G'}}(y \mid x)}}
\newcommand{\Gbare}{\textcolor[RGB]{30,100,200}{\boldsymbol{p_G}}}
\newcommand{\Gprimebare}{\textcolor[RGB]{30,100,200}{\boldsymbol{p_{G'}}}}

\newcommand{\FCPAname}{Frequency-corrected Learning of Ordered Rank Alignment}
\newcommand{\FCPA}{FLORA}

\title{Improving LLMs via Validator-to-Generator Alignment}

\author{
  Juan Diego Rodriguez$^\spadesuit$ \hspace{1em}
  Jocelyn Zhang$^{\spadesuit}$ \hspace{1em}
  Katrin Erk$^{\diamondsuit}$ \hspace{1em}
  Greg Durrett$^\clubsuit$ \\
  \\
  $^\spadesuit$~Department of Computer Science, The University of Texas at Austin \\
  $^\diamondsuit$~Departments of Linguistics and Computer Science, University of Massachusetts Amherst \\
  $^\clubsuit$~Department of Computer Science \& Center for Data Science, New York University \\
  \small \texttt{\{juand-r, jocelynzhang\}@utexas.edu} \quad \texttt{kerk@umass.edu} \quad \texttt{gdurrett@nyu.edu}
}

\begin{document}

\maketitle

\begin{abstract}
Large %
language models are inconsistent: varying prompts or including unrelated information can lead to unexpected changes in model outputs. The generator-validator (G-V) gap is one manifestation of this phenomenon, where LLMs generate responses that they then deem as invalid if re-queried to validate them. %
In this work, we introduce a new formulation of G-V consistency that involves a principled correction for utterance frequency. %
Specifically, generators often assign low likelihood to valid strings simply because those strings are a priori unlikely, which makes naive notions of G-V consistency unworkable. We show that under a natural model of rational agents answering questions with multiple answers, consistency of the validator with a frequency-corrected generator score emerges naturally. 
Our method, \emph{\FCPAname} (\FCPA), is a training objective implementing frequency-corrected G-V consistency for real-world LLMs. Our experimental results show that training with \FCPA{} substantially improves both G-V consistency and generator performance over prior methods, with gains of up to $+27$pp in Pearson correlation on IFEval and HumanEval, while preserving validator quality across all evaluated tasks.\footnote{Code and data available at \url{https://github.com/juand-r/flora}}

\end{abstract}

\section{Introduction}
\label{sec:intro}

Large %
language models (LLMs) can be used in two complementary modes: as \emph{generators} that produce candidate responses, and as \emph{validators} that assess response correctness or felicity. These two modes are crucial for usages of LLMs such as self-refinement \citep{madaan2023selfrefine} and backtracking during chain-of-thought \citep{yao-tree-of-thoughts}. However, these two modes exhibit divergent behavior, reflecting an underlying inconsistency: even frontier LLMs may generate responses with high probability but judge them to be incorrect, or vice-versa. Accordingly, past work has examined training LMs to be explicitly generator-validator (G-V) consistent in the hope that this will also make them more accurate \citep{west2024generative, li-benchmarking2024, rodriguez2025rankalign}. But formalizing this turns out to be surprisingly difficult. Past approaches close the gap on G-V correlation \citep{rodriguez2025rankalign}, but these approaches risk contributing to pathological behaviors like suppression of correct responses or contributing to further inconsistency through contradictory training signals.%

\begin{figure}[t!]
    \centering
    \includegraphics[width=1.0\columnwidth,trim=0mm 50mm 115mm 20mm, %
    clip
    ]{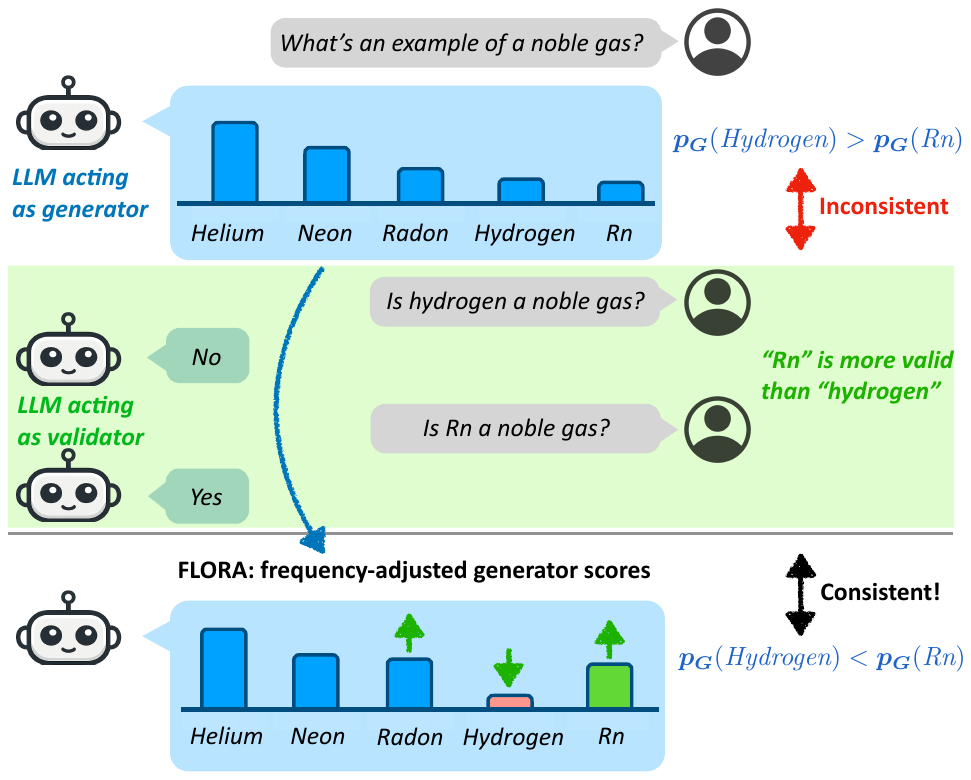}
    \caption{An LLM may generate outputs inconsistent with how it validates them, since low probability options may simply be unlikely to say. FLORA provides a principled correction for generator and allows for training of G-V consistent models.}
    \label{fig:flora-intro}
\end{figure}

This work aims to improve G-V consistency %
through two contributions. First, we axiomatize G-V consistency based on a model of a rational probabilistic agent responding to prompts. Although LLMs are \emph{not} rational agents, this model allows us to examine what the relationship between a generator and a validator should be in cases where the generator may prefer some possible responses over others, but a validator finds them all to be likely. %
We derive a theoretical relationship between generator and validator probabilities, which implies adjusting generator scores by subtracting off \emph{incorrect probabilities}, or how likely a response is to be sampled when an \emph{incorrect} %
response is requested. Since this value reflects the base likelihood of the response, we call the final quantity the \emph{frequency-corrected generator score}. %

\begin{figure}[t!]
    \centering
    \includegraphics[width=0.99\columnwidth, trim=15mm 6mm 210mm 11mm, clip]{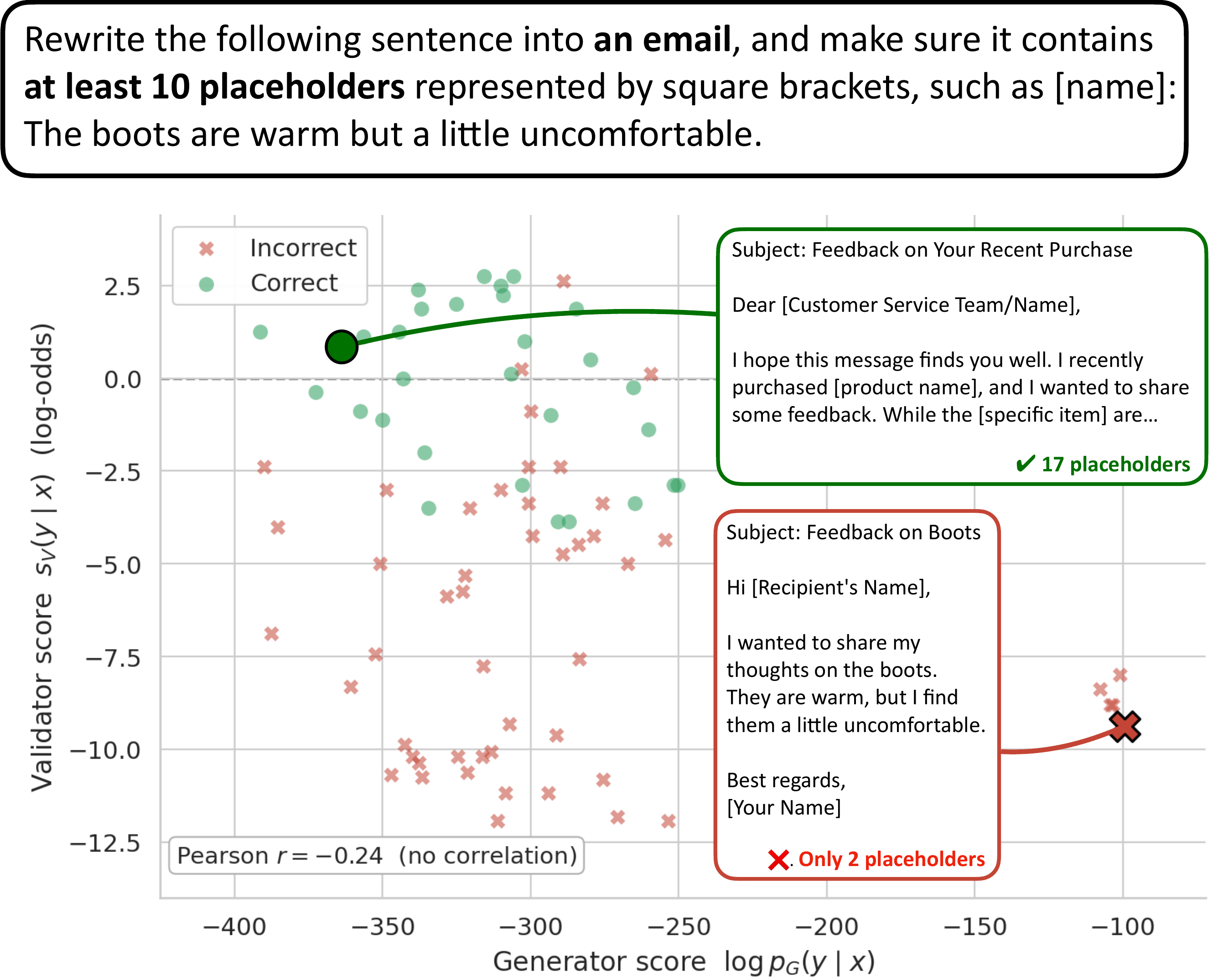}
    \caption{Generator-validator gap on a long-form generation task from IFEval. A short email with few placeholders (red) does not follow the instruction and fails validation but has high generator score. Another email (green) has lower generator score but correctly follows the instruction and is validated as correct. %
    }
    \label{fig:chess_gv_bars}
\end{figure}

Second, we use this relationship in a training objective called %
\textbf{\FCPAname} (\FCPA). This objective encourages rank-alignment of validator scores with frequency-corrected generator scores, representing an approximation of our theoretical G-V consistency for rational agents. %
We fine-tune LLMs using this objective function in combination with standard losses to ensure generator and validator correctness. %

Our experiments focus on tasks where validators outperform generators. 
Our goal is to observe that the generator improves without degradation of the validator. We particularly focus on \emph{generator AUROC}, or ensuring that the frequency-corrected generator distribution correctly ranks the set of positive responses above the set of negative responses, even down into the tail of the distribution, unlike past work which focuses on the head \citep{li-benchmarking2024}. This is a strong check of consistency and of generator correctness. %

We evaluate on three tasks: instruction-following, coding, and eliciting taxonomic knowledge. Two of these tasks are long-form (several sentences, for IFEval, or Python code), unlike previous work which focused on short answers of one or a couple words \citep{rodriguez2025rankalign, li-benchmarking2024} %
Across these tasks and three LLMs, we find that our method closes the G-V gap more strongly than previous work, and leads to improved discriminability of correct and incorrect completions. Concretely, \FCPA{} improves generator AUROC by up to $+7.3$pp and generator-validator correlation by up to $+27$pp over the strongest prior alignment method, while preserving or improving validator quality. %

\section{Background and Motivation}

The extent %
to which LLMs have consistent internal processing is an important scientific question \citep{positionconsistency2026}, and the mismatch between generation and validation is an important kind of inconsistency to measure and repair. Much of what a model ``knows'' never shows up in its sampled outputs, because it lives in discriminative judgments rather than in fluent continuations~\citep{gekhman2025insideouthiddenfactualknowledge}; this is problematic for evaluations targeting model knowledge \citep{wang2024my,biderman2024lessons}. %
Better aligning generators with an LLM's own validation capability is also important for %
reward modeling \citep{yuan2024selfrewarding} and reranking applications %
such as mathematical reasoning \citep{cobbe2021verifiers,lightman2023let}, code generation \citep{chen2021evaluating}, and factual question answering.

\citet{rodriguez2025rankalign} assert that the log-odds of the generator and validator should correlate. However, there are cases where this relationship cannot hold, arising from two main sources: (a) the role of response frequency and (b) aleatoric uncertainty (multiple correct responses).

\paragraph{Motivating Example} A well-known issue with using LLMs to generate responses is surface form ambiguity \cite{holtzman2021sfc}. Given the question \emph{``What do you call it when blood flow to the heart is suddenly blocked?''}, \emph{heart attack} and \emph{myocardial infarction} are both correct answers. A good validator should score them both highly, yet it would be odd to expect a generator to score both highly, since \emph{heart attack} is the more common expression. 

This becomes more complex when questions have multiple possible correct answers beyond paraphrases. Consider a question that admits several valid answers, e.g., \emph{``What's an example of a noble gas?''} (Figure~\ref{fig:flora-intro}). There are 7 possible correct answers to this question, plus potential different surface forms of elements (e.g., \emph{``Helium''} vs.~ \emph{``He''}). It is unlikely that a generator will actually generate \emph{Helium} and \emph{Oganesson} with equal probability when given this question, despite both being correct. Generator scores conflate the correctness of an utterance in response to a prompt with the \emph{frequency} of that utterance. %

\begin{figure*}[t!]
    \centering
    \includegraphics[width=1.0\textwidth,trim=0mm 120mm 0mm 20mm, %
    clip
    ]{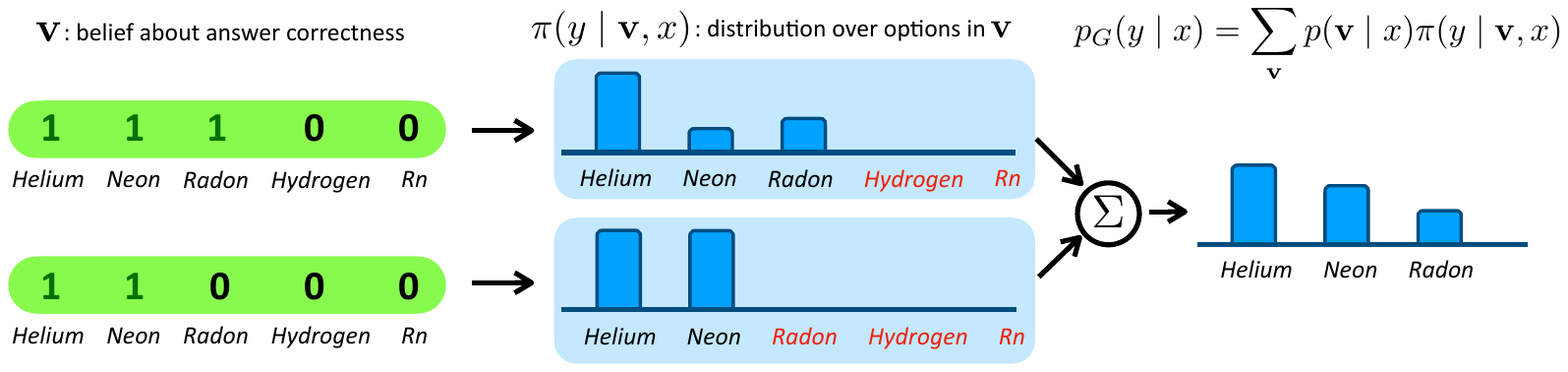}
    \caption{Model of agent response generation. We represent a generator $\Gbare$ as a mixture distribution over latent validity vectors $\mathbf{v}$, which indicate the subset of responses a model believes to be the correct response. $\pi$ places a distribution over $y$ conditioned on each $\mathbf{v}$.}
    \label{fig:flora-pi}
\end{figure*}

There is no single ``perfect'' generator in this setting. One view is that a model should tend towards %
a uniform distribution over possible options \cite{zhang2024forcingdiffusedistributionslanguage}, while another view is that the model’s distribution over valid answers should match the estimated corpus-frequency distribution \citep{timtomovillusion2025}.  %
A perfect validator, in contrast, should certainly say ``Yes'' with probability near 1 to each correct option and ``No'' to incorrect options. 

Figure~\ref{fig:chess_gv_bars} shows this effect in practice on a long-form generation task from IFEval. The validator is imperfect, but most correct responses (green) outscore most incorrect responses (red) on the validator log-odds (y-axis). However, there are incorrect responses that have much higher generator score than correct responses, partially due to being short and generic and sticking close to the prompt.

This example shows that generator and validator scores should not generally be expected to match, or even to correlate linearly, without controlling for competition and frequency effects. 
Next we derive a relationship between them which accounts for both frequency effects and aleatoric uncertainty.

\section{Problem Formulation}
\label{problem_formulation}

\paragraph{Problem Setup} We assume a prompt $x$ and a set of possible responses $\mathcal{Y}$.\footnote{Our analysis will require $\mathcal{Y}$ to be finite in size, but it can be very large: $\Sigma^N$ for a token vocabulary $\Sigma$ and some $N$. We do not assume that $\mathcal{Y}$ is tractable.} For each $y \in \mathcal{Y}$, we assume that there is a validity label $v(x, y) \in \{0, 1\}$. We assume that validity is binary and unambiguous. We consider an agent, which is an LLM for the purposes of this work, operating in two modes: a generator $\G$, and a validator $\V$. For an LLM, $p_V$ and $p_G$ are implemented with different prompts.

\paragraph{Properties of Consistency} We would like to choose $p_V$ and $p_G$ to be consistent, under the intuition that improving consistency will also improve accuracy. %
We can straightforwardly understand $p_V$ as the LM's current ``belief'' that $y$ is a correct response to $x$. However, since there are in general many correct responses, $p_G$ additionally has to model competition between these to act as a generator.

We define $\mathbf{v} = \{0,1\}^{|\mathcal{Y}|}$ as a vector of validities associated with each possible response string, assuming that every possible response is either correct or incorrect, even though the agent may be uncertain as to which.  %
Let $\mathbf{v}_y \in \{0,1\}$ denote the validity assigned to a particular string $y$. In the context of Figure~\ref{fig:flora-intro}, a correct %
$\mathbf{v}$ should assign 1 to the 7 noble gases and all ways of writing them down and 0 to all other strings. Figure~\ref{fig:flora-pi} shows possible $\mathbf{v}$ values in green.

Assume that for a prompt $x$, we have a known and fixed $\mathbf{v}$. We define an agent to be \emph{consistent} with its beliefs $\mathbf{v}$ if (1) its validator returns exactly the responses in $\mathbf{v}$; (2) its generator only places mass on valid responses: $\G > 0$ iff $\mathbf{v}_y = 1$. We call this a \textbf{support constraint} motivated by the maxim of quality \citep{grice1975logic}: conditional on its latent belief state $\mathbf{v}$, the agent never generates a response it believes to be incorrect. 
However, some options (e.g., xenon and krypton) may be assigned very low probability.

In practice, agents have uncertainty over $\mathbf{v}$. We model this as a distribution $p(\mathbf{v} \mid x) > 0$ reflecting the agent's belief over possible response sets given $x$. Like in real LLMs, we assume that every $\mathbf{v}$ has nonzero (but possibly tiny) probability mass. (Note that this distribution is a theoretical construct and cannot be materialized in practice.) We can now define \emph{consistency} with respect to this distribution $p(\mathbf{v} \mid x)$. A validator is consistent if it satisfies the following relationship:
\begin{equation}\label{eq:v-definition}
\Vother := p(v_y = 1 \mid x)
= \sum_{\mathbf{v} : v_y = 1} p(\mathbf{v} \mid x)
\end{equation}

We represent generation through the use of a distribution $\pi(y \mid \mathbf{v}, x)$ that conditions on $\mathbf{v}$.  $\pi$, although intractable to fully represent in practice, effectively describes how mass should be distributed among the correct options of $\mathbf{v}$. To obey the support constraint, $\pi$ only assigns nonzero probability to items $i$ where $\mathbf{v}_i = 1$. But within these items, $\pi$ governs which surface form to use for a given meaning, splitting probability over surface form realizations \citep{zhang2024forcingdiffusedistributionslanguage, holtzman2021sfc, zhang2025noveltybench}, and which response to express when multiple are deemed correct. We define the generator as a mixture model over the $\mathbf{v}$:
\begin{equation}\label{eq:g-definition}
\begin{aligned}
\G &:= \sum_{\mathbf{v}} p(\mathbf{v} \mid x) \, \pi(y \mid \mathbf{v}, x) \\
        &\phantom{:}= \sum_{\mathbf{v}:v_y=1} p(\mathbf{v} \mid x) \, \pi(y \mid \mathbf{v}, x)
\end{aligned}
\end{equation}
This process is depicted in Figure~\ref{fig:flora-pi}.

\subsection{Deriving a G-V relationship}
\label{sec:deriving}

We can now relate the generator and the validator we have defined so far.

We first define one more quantity $\pi'(y \mid \mathbf{v}, x)$. This is a policy for selecting a response when asked for an \emph{incorrect} one, with the related constraint that it never generates an answer it believes to be correct, i.e., $\pi'(y \mid \mathbf{v}, x) = 0$ when $v_a = 1$. From that, we can define
\begin{equation}\label{eq:g-prime-definition}
\begin{aligned}
\Gprime &:= \sum_{\mathbf{v}} P(\mathbf{v} \mid x) \, \pi'(y \mid \mathbf{v}, x) \\
         &\phantom{:}= \sum_{\mathbf{v}:v_y=0} p(\mathbf{v} \mid x) \, \pi'(y \mid \mathbf{v}, x)
\end{aligned}
\end{equation}

\begin{theorem}[Main]\label{thm:main} Assume a problem $x$, solution space $\mathcal{Y}$, and generator $\Gbare$ and validator $\Vbare$ as defined previously. Further assume that ${0 < \Vother < 1}$ for all $y \in \mathcal{Y}$. Then
\begin{equation}\label{eq:main}
    \frac{\Vother}{1 - \Vother} = \frac{\G}{\Gprime} \cdot r(y,x),
\end{equation}
where
\begin{equation*}
    r(y,x) := \frac{\mathbb{E}[\pi'(y \mid \mathbf{v}, x) \mid v_y = 0, x]}{\mathbb{E}[\pi(y \mid \mathbf{v}, x) \mid v_y = 1, x]}.
\end{equation*}
\end{theorem}

\paragraph{Proof Sketch} We give intuition for the proof and a full proof in Appendix~\ref{app:proof}.

The conditional expectation of the policy for picking a correct response from among the set of correct answers given that $a$ is valid is:

\begin{equation}\label{eq:pi-expectation}
\begin{aligned}
&\mathbb{E}\big[\pi(y \mid \mathbf{v}, x) \mid v_y=1, x \big] \\
&\quad = \frac{\sum_{\mathbf{v}: v_y = 1} P(\mathbf{v} \mid x)\, \pi(y \mid \mathbf{v}, x)}{P(v_y = 1 \mid x)} \\
&\quad = \frac{\G}{\Vother} 
\end{aligned}
\end{equation}

Similarly,
\begin{equation}\label{eq:pi-prime-expectation}
\begin{aligned}
&\mathbb{E}\big[\pi'(y \mid \mathbf{v}, x) \mid v_y=0, x \big] \\
&\quad = \frac{\sum_{\mathbf{v}: v_y = 0} p(\mathbf{v} \mid x)\, \pi'(y \mid \mathbf{v}, x)}{P(v_y = 0 \mid x)} \\
&\quad = \frac{\Gprime}{1- \Vother} 
\end{aligned}
\end{equation}

Since $0<\Vother<1$ the main result follows from Eqn \ref{eq:pi-expectation} and \ref{eq:pi-prime-expectation}. 

\paragraph{Intuition} First, we note that $0<\Vother<1$ is a mild condition in practice. We might reasonably expect that an agent backed by a Transformer produces a distribution $p(\mathbf{v} \mid x)$ with support everywhere due to the nature of softmax; no set of answers is ruled out structurally. Our needed condition follows from this.

In the limit of zero epistemic uncertainty, these probabilities become very small and $p$ places all probability mass on one configuration $\mathbf{v}^{\star} \in \{0,1\}^{|A|}$. In this case, the validator probability $\Vother$ will approach 0 or 1, and the generator probabilities become the policies at configuration $\mathbf{v}^{\star}$: $\G =  \pi(y \mid \mathbf{v}^{\star}, x)$ and $\Gprime =  \pi'(y \mid \mathbf{v}^{\star}, x)$. %

Finally, the value of $r(y,x)$ is very important to reasoning about the generator-validator relationship. The numerator of $r$ is the expected value of the probability of a response being generated as a \emph{incorrect} response conditioned on it being marked as incorrect in $\mathbf{v}$. This involves marginalizing over all possible configurations of $\mathbf{v}$ where $v_y = 0$. We can think of this as saying: in aggregate, what fraction of the $\pi'$ mass is assigned to $y$ when it's a valid option? Intuitively, this is related to how frequent we are to say $y$: a string that's simply more frequent will be uttered with higher probability in a larger number of contexts. The denominator of $r$ is similar, but for the case of a correct response.

\subsection{Mapping Rational Agent Behavior to LLM Behavior}

LLMs are not rational agents of the sort in the previous section. However, we can nevertheless seek to impose the regularity we derived in Equation~\ref{eq:main} to their behavior if we can approximate $\Vbare$, $\Gbare$, and $\Gprimebare$. To do so, $\Vother$, $\G$ and $\Gprime$ can be elicited from an LLM as follows. Given template function $T_V: x,y \rightarrow \text{prompt}_V(x,y)$
mapping questions and answers to prompts, %
and template functions $T_G: \q \rightarrow \text{prompt}_G(\q)$ and $T'_G: \q \rightarrow \text{prompt}_G'(\q)$, asking for correct and incorrect answers to a given question, respectively:
\begin{align*}
\Vother   &\approx P_{\mathrm{LLM}}\bigl(\text{Yes} \mid T_V(x,y)\bigr), \\
\G   &\approx P_{\mathrm{LLM}}\bigl(y \mid T_G(x)\bigr),           \\
\Gprime &\approx P_{\mathrm{LLM}}\bigl(y \mid T'_G(x)\bigr).
\end{align*}

There are two sources of error in these approximations. First, LLMs do not have an internally consistent belief $p(\mathbf{v} \mid x)$ about the answers to a problem. Second, even if they did, these prompts do not necessarily elicit rational responses to these.

Nevertheless, we argue that these approximations are useful. Empirically, most of the mass for LLMs when prompted for Yes/No answers lies on the \emph{Yes} and \emph{No} tokens, so at least LLMs can follow these instructions.%
\footnote{In our experiments we aggregate over the tokens \texttt{Yes}, \texttt{yes}, \texttt{YES}, \texttt{\textvisiblespace Yes} and \texttt{\textvisiblespace yes} for the positive class, and \texttt{No}, \texttt{no}, \texttt{NO}, \texttt{\textvisiblespace No} and \texttt{\textvisiblespace no} for the negative class.}. $\G$ follows closely from how LLMs are trained to answer questions or follow instructions. On the other hand, $\Gprime$ is likely less well-approximated, since asking for incorrect responses is not where post-training focuses its efforts; the fidelity of this proxy is an empirical question, especially for smaller or non-instruction-tuned models. %

Letting %
$s_V(y \mid x) := \log \frac{\Vother}{1-\Vother}$ be the validator log-odds, we then have
\begin{equation}
\begin{split}
    s_V(y \mid x) = {}& \log \G - \log \Gprime \\
                      & {} + \log r(y \mid x)
\end{split}
\label{eq:main-2}
\end{equation}

Unfortunately, $\log r(y \mid x)$ depends on conditional expectations over quantities that we cannot directly observe. Intuitively, %
what this term measures is the ratio of the probability of generating $y$ when prompted for a wrong response (and the model thinks $y$ is incorrect) and the probability of generating $y$ when prompted for a correct response (and the model thinks $y$ is correct). Assuming that $\pi$ and $\pi'$ are both capturing the frequency of $y$, and this frequency is not biased by response correctness, %
this ratio can be approximated as 1.

As a result, disregarding $r$, Eqn \ref{eq:main-2} motivates the following adjusted generator score: %
\begin{equation}
    s_{\text{Adj}}(y \mid x) \;:=\; \log \G - \log \Gprime.
    \label{eq:adj-score}
\end{equation}
By construction, $s_{\text{Adj}}$ tracks the validator log-odds $s_V$ up to
the residual $\log r$, and so should correlate better with $s_V$ than the
raw generator score $\log p_G$ does. 
In the next section, we operationalize $s_{\text{Adj}}$ as a training objective for our model.

\section{Training for Frequency Correction}
\label{sec:method}

\paragraph{Data Condition} 
We consider a training setting where we have a dataset $\mathcal{D} = \{(x_i,\{(y_{ij},v_{ij})\})\}$. That is, each $x_i$ is paired with a set of outputs $\{(y_{ij},v_{ij})\}$ where the $y_{ij}$ are the outputs themselves and the $v_{ij}$ are correctness labels. We operate in a semi-supervised setting where $v_{ij}$ may not be available for all $(x_i,y_{ij})$ pairs.

As our training involves imposing consistency, we operate over pairs of points (e.g., similar to \emph{hydrogen} and \emph{Rn} in Figure~\ref{fig:flora-intro}). We define a labeled set $\mathcal{T} = \{((x_i,y_i,v_i),(x_j,y_j,v_j))\}$ of pairs with veracity labels. We define an unlabeled set $\mathcal{U} = \{((x_i,y_i),(x_j,y_j))\}$ of pairs without labels.

Furthermore, throughout this work we require $x_i = x_j$; we sample points \emph{within the same prompt}. We also require $|p_V(y_j \mid x_j) - p_V(y_i \mid x_i)| > \delta$ to avoid noise from too-close validator pairs, following past work \citep{gekhman2025insideouthiddenfactualknowledge, rodriguez2025rankalign}.

\paragraph{Training Objective} We use a loss designed to do two things: (1) teach the generator to rank pairs in the same way as the validator, and (2) use the labeled examples to improve the discriminability of generator and validator scores. %
We train using the following loss, 
where we let $y_j = y^+$ and $y_i = y^-$ denote the higher- and lower-ranked completions under the validator:
\begin{align}
    \mathcal{L} =& \mathcal{L}_{\text{pref}}(x_i,y_i,x_j,y_j) + \notag \\
    &\lambda_G  (v_i \mathcal{L}_\text{NLL}(y_i \mid x_i) + v_j \mathcal{L}_\text{NLL}(y_j \mid x_j)) + \notag \\
    & \lambda_V  (\mathcal{L}_\text{NLL}(v_i \mid x_i, y_i) + \mathcal{L}_\text{NLL}(v_j \mid x_j, y_j))
\label{eq:loss}
\end{align}

\noindent The $\mathcal{L}_{\text{pref}}$ loss is always active, while $\mathcal{L}_\text{NLL}$ terms are only active for labeled examples. The generator terms only apply to positive examples (i.e., $v_i = 1$ or $v_j = 1$) and the validator terms are only present %
for labeled examples, which have ground truth Yes/No completions. %

\paragraph{Preference Loss}

We use a pairwise logistic preference loss given by 
\begin{equation}
\small 
    \mathcal{L}_{\text{pref}}(x, y^-, y^+) = -\log \sigma\!\left( s(y^+ | x) - s(y^- | x) \right)
\end{equation}
where $\sigma$ is the logistic sigmoid and $s_G(y \mid x)$ is the generator's log-likelihood of completion $y$ given prompt $x$,
\begin{equation}
    s(y \mid x) = \sum_{t=1}^{|y|} \log P_{\text{LLM}}(y_t \mid x, y_{<t}).
\end{equation}

To account for completion frequency, we consider two empirical estimators
of the adjusted score $s_{\text{Adj}}$ from Eqn~\ref{eq:adj-score}
(\S\ref{sec:deriving}). The first, \textbf{NegFC}, directly substitutes a negative-prompt
elicitation for $p_{G'}$:
\begin{equation}
s_{\text{Neg}}(y \mid x) = s_G(y \mid x) - \log P_{\text{LLM}}(y \mid T'_G(x)).
\end{equation}
The second, \textbf{Unconditional FC}, uses the unconditional log probability~\citep{meister2022typical}
as a cheaper proxy,
\begin{equation}
    s_{\text{PMI}}(y \mid x) = s_G(y \mid x) - \log P_{\text{LLM}}(y),
\end{equation}
which captures a similar frequency-penalization effect without requiring an
elicitation prompt $T'_G$ for incorrect answers.

\paragraph{Sampling strategy} Finally, we ensure that the preference losses are consistent with the labels. For any pair $(x,y^-), (x,y^+)$ where $s_V(y^+ \mid x) > s_V(y^- | x) + \delta $, the preference loss will push the generator score of $(x,y^-)$ down and $(x,y^+)$ up; this is incorrect if $y^-$ is a positive example or $y^+$ is a negative example. In addition, if $y^-$ is positive the preference loss and generator $\mathcal{L}_\text{NLL}$ will compete in opposite directions. %
To avoid these problems, we only keep a labeled $(x,y^-)$ when it is negative, and similarly we only keep a labeled $(x,y^+)$ when it is positive.

\section{Experimental Setup}
\label{sec:experiments}

\subsection{Tasks and Datasets}

We evaluate on three datasets which we adapted from previous work to test our method on instruction following, coding, and conceptual knowledge.  %
We use IFEval \citep{zhou2023ifeval} for instruction following, HumanEval \cite{chen2021evaluating} for Python coding, and a mix of datasets to probe for knowledge of taxonomic category relations (hyponymy)  \citep{rosch1975cognitive, banks2023category, stoinski2024thingsplus, castro2021category, van2004category, uyeda1980prototypicality}.
Each dataset has a different notion of correctness, detailed below.
These datasets are described here, with further details on datasets given in Appendix \ref{app:dataset-details}. Prompt templates for each of these tasks are shown in Appendix \ref{app:prompts}.

\paragraph{Instruction Following} %
This task uses prompts from IFEval~\citep{zhou2023ifeval}, e.g., ``\emph{Write a poem that's at least 350 words about the beauty of eucalyptus trees and their many uses.}'' %
which specify explicit content and format constraints. A correct response should follow all constraints in the prompt, while an incorrect response violates one or more. We generate, using GPT-4o mini, correct responses via paraphrases of the original prompt and incorrect responses by introducing violations (e.g., missing required elements or breaking constraints). Labels are assigned using rule-based checks for verifiable constraints together with LLM-as-a-judge evaluations for content correctness where deterministic verification is not possible. 
We use 79 prompts for the training set and 20 for the test set.

\paragraph{Coding} This task uses prompts from the HumanEval \citep{chen2021evaluating} benchmark of Python programming exercises. Each prompt asks to complete a Python function to solve a given problem. Here correct responses are those which pass all the unit tests, while incorrect ones fail one or more. We use various LLMs to generate correct and incorrect responses. %
To stress-test our method with examples that are correct but unlikely, we systematically convert the variables of all the correct solutions to uppercase.

\paragraph{Hyponymy} This task consists of giving exemplars to categories, i.e., answering ``An example of a X is a \_\_'' for various nouns X. 
We evaluated on the ten categories used in the original experiments of \citep{rosch1975cognitive}, supplemented with GPT-5-generated incorrect completions in order to obtain a balanced test set. We decided on a cutoff between items that are members of the category while those that are not. While there is some subjectivity in this decision, our cutoff is validated by the fact that LLMs with validator prompts can successfully distinguish between the positive and negative classes (Table \ref{tab:main-results-multi-val}). \newline

For all three tasks, we evaluate on a held-out set of prompts. For IFEval and HumanEval, we randomly split the data. For Hyponymy we train on a different set of categories than the ten in \citep{rosch1975cognitive}. Further details of the dataset composition and construction process is given in Appendix \ref{app:dataset-details}.

\subsection{Target Models for G-V Alignment}

As we defined the dataset at the beginning of Section~\ref{sec:method}, our target models for G-V consistency do not necessarily have to be those from which responses were generated. In fact, having a range of responses (as we generated for each dataset) that we require G-V consistency on allows us to stress-test each model beyond the mode of the distribution that it generates.

We use a range of models including those in the Gemma family, namely Gemma-2-9b-it (\textbf{G2-9b-it}) and Gemma-4-31b-it (\textbf{G4-31b-it}), and Qwen family, namely Qwen-3.5-9B (\textbf{Q3.5-9b}). For each task, we select models in an appropriate performance range: for instance, on IFEval, we only use larger instruction-tuned models because smaller models are generally not capable enough at the task. For each task and model combination, we only proceed with models where validator AUROC is over 65\%); this eliminates Gemma-2-9b-it from the HumanEval experiments.   %

\subsection{Evaluation}

For all methods, we evaluate both using the raw generator scores $s_G(y \mid x) = \log P_{LLM}(y \mid x)$ as well as the two corrected forms: frequency-corrected scores $s_{\text{PMI}}= s_G(y \mid \q) - \log P_{LLM}(y)$, and negative prompting-corrected scores $s_{\text{Neg}}= s_G(y \mid \q) - s_G'(y\mid \q)$.
All metrics are computed independently for each prompt, and averaged over prompts. %

\paragraph{Accuracy} We %
evaluate the validator and generator performance at discriminating correct and incorrect answers through AUROC scores of the associated classifiers: $\G > \tau_G$ and $s_V(y \mid x) > \tau_V$ for thresholds $\tau_G$ and $\tau_V$. Each defines an ROC curve and we refer to the corresponding AUROC scores as $\mathbf{ROC_G}$ and $\mathbf{ROC_V}$. For the generator, we use one of several different possible generator scores: $\Gbare$, $s_{\text{PMI}}$ 
or $s_{\text{Neg}}$. We additionally measure the \textbf{validator accuracy} at $\tau_V =0$ in order to measure how well-calibrated the validator is. %

We focus primarily on $\mathbf{ROC_G}$; our focus in this work is on tasks where the validator is stronger than the generator and can be used to improve the generator's performance.

\paragraph{Correlations} Following \citet{rodriguez2025rankalign}, we measure the Pearson correlation $\rho$ between the validator and (possibly-corrected) generator scores  over the full set of candidate answers. %

\begin{table*}[t]
\centering
\small
\setlength{\tabcolsep}{6pt}
\begin{tabular}{l cc cc cc}
\toprule
& \multicolumn{2}{c}{\textbf{IFEval}} & \multicolumn{2}{c}{\textbf{HumanEval}} & \multicolumn{2}{c}{\textbf{Hyponymy}}  \\
\cmidrule(lr){2-3} \cmidrule(lr){4-5} \cmidrule(lr){6-7}
\textbf{Eval-time score} & ROC$_G$ & $\rho$ & ROC$_G$ & $\rho$ & ROC$_G$ & $\rho$ \\
\midrule
$s_G$ (raw)                & 57.5 ± 1.9     & 16.3 ± 3.2     & 65.7 ± 1.5     & 34.1 ± 2.2     & 66.9 ± 2.6     & 31.5 ± 2.6     \\
$s_{\text{PMI}}$ (Uncond. FC)  & \B{81.0} ± 1.2 & \B{44.0} ± 2.8 & \B{72.9} ± 1.3 & 22.4 ± 3.0     & 77.7 ± 2.2     & \B{50.0} ± 2.7 \\
$s_{\text{Neg}}$ (NegFC)   & 64.5 ± 1.8     & 25.3 ± 3.4     & 71.1 ± 1.6     & \B{38.5} ± 2.5 & \B{81.2} ± 4.0 & 48.0 ± 5.4     \\
\bottomrule
\end{tabular}
\caption{Eval-time frequency correction applied to Gemma-2-9b-it, except HumanEval, which uses Gemma-4-31b-it. Both NegFC and Unconditional FC improve over the raw generator score, but neither dominates across all tasks. Subsequent sections apply the per-task best eval-time correction to all methods for a fair comparison.}
\label{tab:eval-time}
\end{table*}

\subsection{Baselines}

We compare our method against the following baselines in addition to the \textbf{Base} (untrained) model: 

\paragraph{SFT} We sample pairs of completions as in RankAlign, but use a Supervised Fine-Tuning (SFT) loss, i.e., only the negative log-likelihood terms from Eqn. \ref{eq:loss}.

\paragraph{Consistency FT} \citet{li-benchmarking2024} SFT only on the subset of examples where generator and validator agree. We use the version from \citet{rodriguez2025rankalign}, which measures agreement via  
$\mathds{1}_{l_G(z, y_A) > t_G} = \mathds{1}_{l_V(z, y_A) > t_V}$
where $t_G$ and $t_V$ are thresholds set as the average generator and validator scores over the dataset.

\paragraph{RankAlign} We use the preference loss of RankAlign \cite{rodriguez2025rankalign}. Compared to our method this (1) does not apply any frequency correction during training; (2) does not enforce sampled pairs to have the same prompt during training, (3) does not use NLL terms.

\section{Results}
\label{sec:results}

\subsection{Frequency Correction Helps at Test Time}
\label{sec:results:eval-time}

\textit{Frequency correction improves generator--validator consistency even before training.}

We first ask whether frequency correction, applied at test time, improves the
generator's ability to discriminate correct from incorrect answers and its consistency with the model's own validator.
Table~\ref{tab:eval-time} compares the raw generator score $s_G$ with the two corrected variants
introduced in Section~\ref{sec:method}: NegTC ($s_{\text{Neg}}$), which subtracts the log probability when using the negative prompt, 
and Unconditional FC ($s_{\text{PMI}}$), which subtracts the unconditional log probability of the completion.

Both corrections improve ROC$_G$ and $\rho$ over the raw generator score on all tasks. This indicates that some amount of apparent G-V gap goes away when using these corrected terms, indicating that they form a better starting point for improving G-V consistency and generator discriminability.

\subsection{Main Results}
\label{sec:results:main}

\textit{FLORA outperforms prior consistency alignment methods on both ROC$_G$ and $\rho$.}

Table~\ref{tab:main-results-multi-nopm} compares methods on ROC$_G$ and $\rho$ for IFEval, HumanEval, and Hyponymy datasets. All methods are evaluated with the per-task best eval-time correction identified in Section~\ref{sec:results:eval-time}. Full results with standard errors are given in Table \ref{tab:main-results-multi} in Appendix \ref{app:additional}.

\begin{table*}[t]
\centering
\footnotesize
\setlength{\tabcolsep}{4pt}
\begin{tabular}{l cc cc cc cc cc cc}
\toprule
& \multicolumn{4}{c}{\textbf{IFEval}} & \multicolumn{4}{c}{\textbf{HumanEval}} & \multicolumn{4}{c}{\textbf{Hyponymy}} \\
\cmidrule(lr){2-5} \cmidrule(lr){6-9} \cmidrule(lr){10-13}
& \multicolumn{2}{c}{G2-9b-it} & \multicolumn{2}{c}{Q3.5-9b} & \multicolumn{2}{c}{G4-31b-it} & \multicolumn{2}{c}{Q3.5-9b} & \multicolumn{2}{c}{G2-9b-it} & \multicolumn{2}{c}{Q3.5-9b} \\
\cmidrule(lr){2-3} \cmidrule(lr){4-5} \cmidrule(lr){6-7} \cmidrule(lr){8-9} \cmidrule(lr){10-11} \cmidrule(lr){12-13}
\textbf{Method} & ROC$_G$ & $\rho$ & ROC$_G$ & $\rho$ & ROC$_G$ & $\rho$ & ROC$_G$ & $\rho$ & ROC$_G$ & $\rho$ & ROC$_G$ & $\rho$ \\
\midrule
Base            & 78.2     & 35.5     & 69.3     & 21.5     & 72.9     & 22.4     & 65.7     & 30.4     & 81.2     & 48.0     & 73.5     & 42.9     \\
SFT             & 74.4     & 29.5     & 73.1     & 24.1     & 84.5     & 28.5     & \B{94.1}     & 56.3     & 88.1     & 64.0     & 82.6     & 48.6     \\
Consistency FT  & 58.7     & 0.8     & 69.8     & 22.4     & 74.9     & 49.0     & 92.6     & 53.5     & 85.3     & 50.8     & 80.1     & 56.6     \\
RankAlign       & 74.9     & 29.9     & 75.4     & 32.8     & 86.9     & 40.9     & 90.2     & \B{69.5}     & 91.6     & \B{75.2}     & 88.2     & \B{70.4}     \\
\midrule
\FCPA{}-PMI    & \B{85.1} & \B{62.8} & \B{79.6}     & 43.3     & 92.2     & 66.8     & 83.2     & 58.9     & \B{92.5} & 73.8     & 87.3     & 69.4     \\
\FCPA{}-Neg     & 60.8     & 34.4     & 66.0     & \B{44.7}     & \B{94.2} & \B{76.3} & 88.9     & 68.5     & \B{92.3} & 72.2     & \B{88.2}     & 69.4     \\
\bottomrule
\end{tabular}
\caption{\textbf{Main results across tasks and models.} ROC$_G$ (generator AUROC) measures generator discriminability, and $\rho$ measures generator--validator Pearson correlation). All methods use the per-task best eval-time correction.}
\label{tab:main-results-multi-nopm}
\end{table*}

\FCPA{} achieves a higher $\text{ROC}_G$ than the other baselines in five 
out of six settings (tying RankAlign on Hyponymy with Qwen-3.5-9B), with
gains ranging from +0.9 to +7.3pp over the next-best method. G-V consistency improves by a wider margin, with $\rho$ +27pp over the next-best method on both IFEval with Gemma-2-9b-it
(\FCPA{}-PMI: 62.8 vs.~35.5 for Base) and HumanEval (\FCPA{}-Neg: 76.3
vs.~49.0 for Consistency FT). On the two Hyponymy settings, \FCPA{} is
competitive on both metrics, trailing RankAlign by at most 1.4pp on $\rho$.

\paragraph{\FCPA{} Preserves Validator Quality}
\label{sec:results:validator}

\textit{Improvements in G-V consistency do not come at the cost of validator quality.}

A standard concern with preference-style training is \emph{likelihood displacement} \citep{razin2025unintentional, xu2024contrastive}: the model may appear better aligned only because its validator has degraded. To rule this out, we measure the validator accuracy and AUROC. %
Table~\ref{tab:main-results-multi-val} shows the validator accuracies per-method ($\text{ROC}_V$ and $\text{Acc}_V$).%

\begin{figure}[t]
    \centering
    \includegraphics[width=\columnwidth,trim=0 0 0 0,clip]{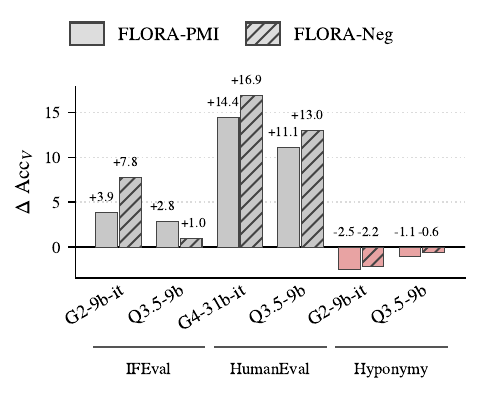}
    \caption{Change in validator accuracy ($\Delta\,\text{Acc}_V$, Method $-$ Base, in percentage points) after \FCPA{} training. \textbf{Our primary aim is for validator accuracy not to degrade after \FCPA{} training.} However, \FCPA{} actually preserves or improves validator accuracy across all five settings, with the largest gains on HumanEval (+14--17pp) and IFEval/G2-9b-it (+4--8pp).} %
    \label{fig:validator-displacement}
\end{figure}

\paragraph{The effect of frequency correction}

Table~\ref{tab:ablations} ablates 
the frequency correction term applied \emph{during training}. Here the ``w/o typ.\ corr.'' variant trains on the raw generator log-likelihood. Both variants apply the same eval-time correction, so the comparison isolates the effect of frequency correction during training. Removing it hurts both ROC$_G$ and $\rho$ on every task, with the largest drops on HumanEval ($-6.3$pp ROC$_G$, $-16.1$pp $\rho$).

\begin{table}[t]
\centering
\small
\setlength{\tabcolsep}{6pt}
\begin{tabular}{l cc cc}
\toprule
& \multicolumn{2}{c}{\textbf{\FCPA{} (full)}} & \multicolumn{2}{c}{\textbf{w/o typ.\ corr.}} \\
\cmidrule(lr){2-3} \cmidrule(lr){4-5}
\textbf{Task} & ROC$_G$ & $\rho$ & ROC$_G$ & $\rho$ \\
\midrule
IFEval     & 85.1 & 62.8 & 78.5 & 61.1 \\
HumanEval  & 94.2 & 76.3 & 87.9 & 60.2 \\
Hyponymy   & 92.5 & 73.8 & 90.9 & 73.1 \\
\bottomrule
\end{tabular}
\caption{Removing frequency correction during training (shown here: Gemma-2-9b-it) has a detrimental effect on both G-V consistency and generator performance. Frequency correction is applied at test time in all cases.}
\label{tab:ablations}
\end{table}

\subsection{Qualitative Analysis of Alignment}

\begin{figure*}[t!]
    \centering
    \includegraphics[width=\textwidth,trim=0 330 0 0,clip]{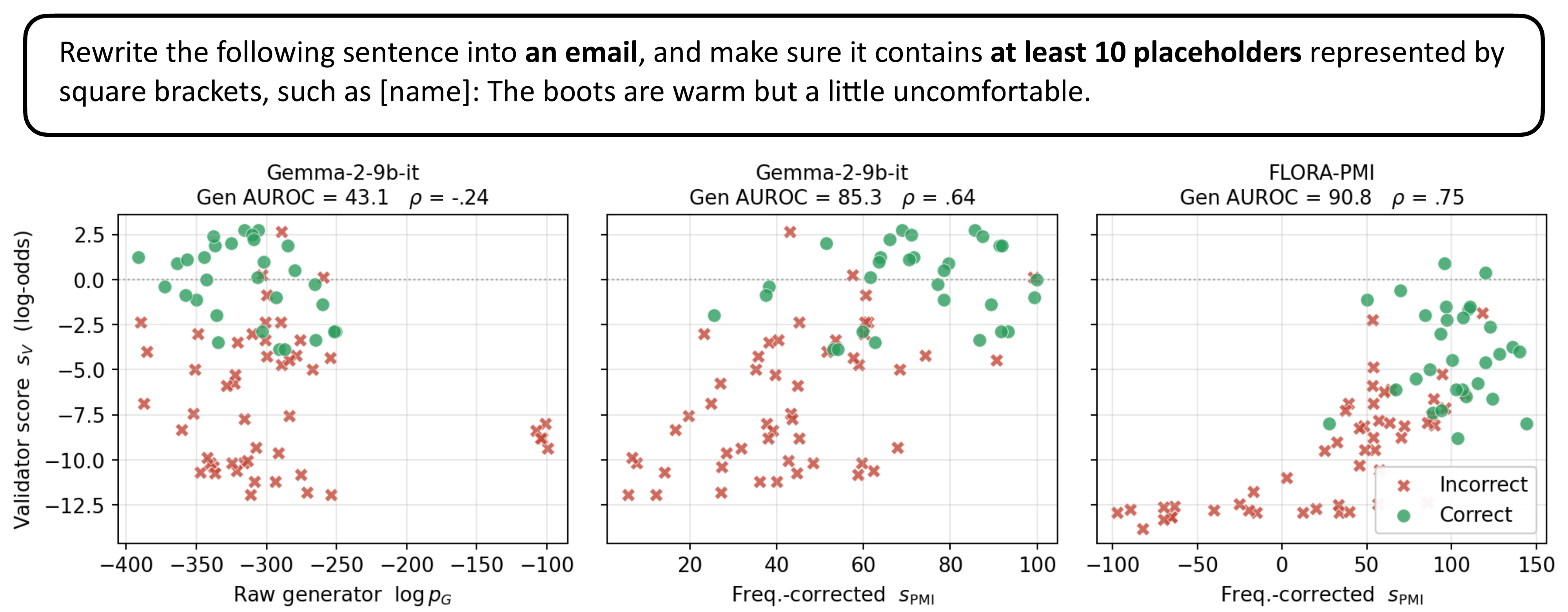}
    \caption{Generator score versus validator log-odds for candidate responses to a single IFEval prompt (base Gemma-2-9b-it; green: correct, red: incorrect). Each panel plots that model's own validator on the $y$-axis. \emph{Left:} the raw generator score $\log p_G$ is uncorrelated with correctness and does not separate correct from incorrect completions (Gen AUROC $43.1$, $\rho{=}{-}.24$). \emph{Middle:} applying the frequency (PMI) correction at test time already recovers most of the alignment (AUROC $85.3$, $\rho{=}.64$). \emph{Right:} after \FCPA{}-PMI training, the corrected generator score ranks correct above incorrect completions and tracks the validator (AUROC $90.8$, $\rho{=}.75$).}
    \label{fig:ifeval_93}
\end{figure*}

Across many prompts, we observe cleaner consistency between generator scores and validator scores after applying our method. Figure~\ref{fig:ifeval_93} illustrates a case of this for a prompt in IFEval and a set of possible completions. X-axes in the left panel are the raw generator scores, while x-axes in the right two panel show the corrected generator scores $s_{\mathrm{PMI}}$.
For the base model (left), the relationship between generator scores and validator scores is noisy: some high-likelihood generations receive low validator scores and vice versa, indicating a misalignment between what the generator prefers and what the validator deems correct. Using frequency correction for the generator scores at test time improves both generator-validator correlation and generator ROC (middle panel).
After training, (right panel) the relationship between generator and validator scores becomes even more structured, with a clearer trend in which higher generator scores correspond to higher validator scores. %

\section{Related Work}
\label{sec:related}

\paragraph{Generator-Validator Gap.} Generator-validator gaps have been discussed in a number of contexts, including \citep{west2024generative}, \citep{li-benchmarking2024}, and \cite{rodriguez2025rankalign}. Our work builds on these notions and is the first to derive a principled relationship between generator probabilities and validator log-odds, which \citep{rodriguez2025rankalign} treated heuristically.

\paragraph{Preference Learning.} The ranking-based preference objective we use is similar to DPO \cite{rafailov2023dpo}. In our case, the validator produces preferences for the generator, broadly related to self-rewarding LMs \cite{yuan2024selfrewarding}. However, rather than a general post-training method, we view FLORA as a mechanism to achieve \emph{consistency} in an LLMs' predictions.

\paragraph{Frequency correction.} Past work has explored similar ideas in frequency correction. \citet{holtzman2021sfc} correct for frequency in multiple-choice settings. Their domain-conditional PMI is similar to our PMI objective, but it is motivated as a heuristic. \citet{liu-etal-2021-dexperts} uses the idea of ``anti-experts'' for controlled text generation, which gives rise to a similar form (subtracting log probabilities from the anti-expert). \citet{li-etal-2023-contrastive} then extend this idea to contrasting strong and weak language models. However, none of these approaches targets consistency per se.

\paragraph{Theoretical Formulation.} We are not the first to analyze LLM behavior by decomposing output distributions over a latent variable. \citet{xie2022explanation} formulate in-context learning as posterior inference over a latent document concept, and \citet{bigelow2025belief} similarly adopt a latent variable formulation to unify in-context learning and activation steering. Our derivation in \S\ref{sec:deriving} uses a similar idea, but applies it to a different question: the discrepancy between a single model's generator and validator distributions on a given prompt.

\section{Conclusion}
\label{sec:conclusion}

In this paper, we presented FLORA, a method for improving LLMs via validator-to-generator alignment. We derive a relationship between generator probabilities and validator log-odds based on a model of how a rational agent might generate a response. This relationship suggests a frequency-based correction factor. When trained to align validator and generator with this factor, we see improved results across taxonomic categorization (Hyponymy) and two long-form generation problems, IFEval and HumanEval.

\section*{Limitations}

One limitation of this work is a gap between the theoretical motivation and practical implementation. We rely on an assumption that the intractable $r$ term in Equation $\ref{eq:main}$ is close to 1. Although we believe this is a reasonable assumption as discussed in the text, it is difficult to validate this as $r$ is intractable to estimate even on relatively simple examples. Nevertheless, we see that our theoretically-derived training objective performs well empirically, lending credence to our method.

A second limitation is the focus on evaluating fixed sets of outputs. We focus on evaluating the tails of the model's distribution: samples which are potentially low likelihood but which we still believe the model should rank correctly. We believe this setting is relevant, as LLM judges are frequently asked to reckon about fixed responses from other models, and various kinds of self-consistency or best-of-N approaches use a related kind of scoring. However, we do not demonstrate an impact on the 1-best answers produced by greedy decoding.

Furthermore, our framework assumes that correctness is a binary notion. Relaxing this to accommodate partially-correct answers would be an interesting direction for future.

Finally, we note that evaluating within-prompt rather than across a diverse set of prompts prevents us from evaluating directly on the datasets used to measure the G-V gap in \citep{rodriguez2025rankalign}, because most prompts in that work only have a small set of completions. To still enable a comparison, we re-implement RankAlign as a baseline and evaluate it on our datasets alongside \FCPA{} (Tables~\ref{tab:main-results-multi}, \ref{tab:main-results-multi-val}).

\section*{Acknowledgments}

We thank Jacob Andreas for insightful discussion around this work. This work was supported by NSF CAREER Award IIS-2145280, NSF grant IIS-2433071, the NSF AI Institute for Foundations of Machine Learning (IFML), and the NSF under Cooperative Agreement 2421782 and the Simons Foundation grant MPS-AI-00010515 awarded to the NSF-Simons AI Institute for Cosmic Origins — CosmicAI, \url{https://www.cosmicai.org/}. We also thank members of the TAUR Lab for helpful feedback.

\bibliography{references}

\appendix

\section{Proof of Theorem~\ref{thm:main}}
\label{app:proof}

\paragraph{Step 1: Rewrite $p_G$ and $p_{G'}$.}
By definition,
\[
\G = \sum_{\mathbf{v}} p(\mathbf{v} \mid x)\,\pi(y \mid \mathbf{v}, x).
\]
Split the sum into $\{\mathbf{v} : v_y = 1\}$ and $\{\mathbf{v} : v_y = 0\}$. By the support constraint on $\pi$, every term in the second sum vanishes, so
\begin{equation}\label{eq:pg-restricted}
\G = \sum_{\mathbf{v} : v_y = 1} p(\mathbf{v} \mid x)\,\pi(y \mid \mathbf{v}, x).
\end{equation}
By the same reasoning,
\begin{equation}\label{eq:pgprime-restricted}
\Gprime = \sum_{\mathbf{v} : v_y = 0} p(\mathbf{v} \mid x)\,\pi'(y \mid \mathbf{v}, x).
\end{equation}

\paragraph{Step 2: Express with conditional expectation.}
For any function $f(\mathbf{v})$ and set $A \subseteq \{0,1\}^{|\mathcal{Y}|}$ with $P(\mathbf{v} \in A) > 0$,
\[
\mathbb{E}\!\left[f(\mathbf{v}) \mid \mathbf{v} \in A\right]
= \frac{\sum_{\mathbf{v} \in A} p(\mathbf{v} \mid x)\,f(\mathbf{v})}{\sum_{\mathbf{v} \in A} p(\mathbf{v} \mid x)}.
\]
Applying this with $f(\mathbf{v}) = \pi(y \mid \mathbf{v}, x)$ and $A = \{\mathbf{v} : v_y = 1\}$, the fact that $\Vother > 0$ ensures that the denominator is greater than 0, as the denominator equals $\Vother$ by~\eqref{eq:v-definition}. The numerator equals $\G$ by~\eqref{eq:pg-restricted}. Hence
\begin{equation}\label{eq:pi-cond-exp}
\mathbb{E}\!\left[\pi(y \mid \mathbf{v}, x) \mid v_y = 1, x\right] = \frac{\G}{\Vother}.
\end{equation}
The analogous calculation with $f(\mathbf{v}) = \pi'(y \mid \mathbf{v}, x)$ and $A = \{\mathbf{v} : v_y = 0\}$, using $\Vother < 1$ to ensure $1 - \Vother > 0$, yields
\begin{equation}\label{eq:pi-prime-cond-exp}
\mathbb{E}\!\left[\pi'(y \mid \mathbf{v}, x) \mid v_y = 0, x\right] = \frac{\Gprime}{1 - \Vother}.
\end{equation}

\paragraph{Step 3: Take the ratio.}
The denominators on the right-hand sides of~\eqref{eq:pi-cond-exp} and~\eqref{eq:pi-prime-cond-exp} are strictly positive by the hypothesis $0 < \Vother < 1$. The numerators are nonnegative as sums of nonnegative terms. %

Dividing~\eqref{eq:pi-prime-cond-exp} by~\eqref{eq:pi-cond-exp},
\[
\frac{\Gprime / (1 - \Vother)}{\G / \Vother}
= \frac{\mathbb{E}\!\left[\pi'(y \mid \mathbf{v}, x) \mid v_y = 0, x\right]}{\mathbb{E}\!\left[\pi(y \mid \mathbf{v}, x) \mid v_y = 1, x\right]}
\]
The left-hand side simplifies to
\[
\frac{\Vother}{1 - \Vother} \cdot \frac{\Gprime}{\G},
\]
and rearranging gives
\[
\frac{\Vother}{1 - \Vother} = \frac{\G}{\Gprime} \cdot r(y, x). \qedhere
\]

\section{Additional Results}
\label{app:additional}

Full results on generator ROC (ROC$_G$) and Pearson correlation ($\rho$) are given in Table \ref{tab:main-results-multi}. Each entry is the average score across prompts, while $\pm$ indicates the standard error.

\begin{table*}[t]
\centering
\footnotesize
\setlength{\tabcolsep}{4pt}
\resizebox{\textwidth}{!}{%
\begin{tabular}{l cc cc cc cc cc cc}
\toprule
& \multicolumn{4}{c}{\textbf{IFEval}} & \multicolumn{4}{c}{\textbf{HumanEval}} & \multicolumn{4}{c}{\textbf{Hyponymy}} \\
\cmidrule(lr){2-5} \cmidrule(lr){6-9} \cmidrule(lr){10-13}
& \multicolumn{2}{c}{G2-9b-it} & \multicolumn{2}{c}{Q3.5-9b} & \multicolumn{2}{c}{G4-31b-it} & \multicolumn{2}{c}{Q3.5-9b} & \multicolumn{2}{c}{G2-9b-it} & \multicolumn{2}{c}{Q3.5-9b} \\
\cmidrule(lr){2-3} \cmidrule(lr){4-5} \cmidrule(lr){6-7} \cmidrule(lr){8-9} \cmidrule(lr){10-11} \cmidrule(lr){12-13}
\textbf{Method} & ROC$_G$ & $\rho$ & ROC$_G$ & $\rho$ & ROC$_G$ & $\rho$ & ROC$_G$ & $\rho$ & ROC$_G$ & $\rho$ & ROC$_G$ & $\rho$ \\
\midrule
Base            & 78.2\stdv{2.2}     & 35.5\stdv{7.5}     & 69.3\stdv{3.7}     & 21.5\stdv{7.8}     & 72.9\stdv{1.3}     & 22.4\stdv{3.0}     & 65.7\stdv{1.6}     & 30.4\stdv{2.4}     & 81.2\stdv{4.0}     & 48.0\stdv{5.4}     & 73.5\stdv{1.1}     & 42.9\stdv{2.5}     \\
SFT             & 74.4\stdv{2.6}     & 29.5\stdv{6.3}     & 73.1\stdv{3.4}     & 24.1\stdv{4.8}     & 84.5\stdv{1.1}     & 28.5\stdv{3.0}     & 94.1\stdv{0.7}     & 56.3\stdv{2.2}     & 88.1\stdv{1.3}     & 64.0\stdv{2.8}     & 82.6\stdv{1.9}     & 48.6\stdv{2.9}     \\
Consistency FT  & 58.7\stdv{3.3}     & 0.8\stdv{7.9}     & 69.8\stdv{3.6}     & 22.4\stdv{6.6}     & 74.9\stdv{1.4}     & 49.0\stdv{2.3}     & 92.6\stdv{0.8}     & 53.5\stdv{2.0}     & 85.3\stdv{1.4}     & 50.8\stdv{3.4}     & 80.1\stdv{1.9}     & 56.6\stdv{2.2}     \\
RankAlign       & 74.9\stdv{3.6}     & 29.9\stdv{7.6}     & 75.4\stdv{3.9}     & 32.8\stdv{7.7}     & 86.9\stdv{1.0}     & 40.9\stdv{3.0}     & 90.2\stdv{1.3}     & 69.5\stdv{1.7}     & 91.6\stdv{1.3}     & \B{75.2}\stdv{2.7}     & 88.2\stdv{1.8}     & \B{70.4}\stdv{3.0}     \\
\midrule
\FCPA{}-PMI    & \B{85.1}\stdv{2.7} & \B{62.8}\stdv{4.4} & \B{79.6}\stdv{3.4}     & 43.3\stdv{7.3}     & 92.2\stdv{0.8}     & 66.8\stdv{1.8}     & 83.2\stdv{1.2}     & 58.9\stdv{2.0}     & \B{92.5}\stdv{1.1} & 73.8\stdv{2.3}     & 87.3\stdv{2.1}     & 69.4\stdv{2.6}     \\
\FCPA{}-Neg     & 60.8\stdv{4.3}     & 34.4\stdv{6.4}     & 66.0\stdv{3.1}     & \B{44.7}\stdv{4.9}     & \B{94.2}\stdv{1.5} & \B{76.3}\stdv{1.3} & 88.9\stdv{1.1}     & 68.5\stdv{1.5}     & \B{92.3}\stdv{1.5} & 72.2\stdv{3.1}     & \B{88.2}\stdv{2.1}     & 69.4\stdv{2.9}     \\
\bottomrule
\end{tabular}%
}
\caption{Main results across tasks and models. ROC$_G$ measures generator discriminability, and $\rho$ measures generator--validator correlation. All methods use the per-task best eval-time correction. Models: G2-9b-it= Gemma-2-9b-it, Q3.5-9b = Qwen-3.5-9B, G4-31b-it = Gemma-4-31b-it.}
\label{tab:main-results-multi}
\end{table*}

The full results on validator accuracy (Acc$_V$) and validator AUROC (ROC$_V$) are given in Table \ref{tab:main-results-multi-val}.

\begin{table*}[t]
    \centering
    \footnotesize
    \setlength{\tabcolsep}{4pt}
    \resizebox{\textwidth}{!}{%
    \begin{tabular}{l cc cc cc cc cc cc}
    \toprule
    & \multicolumn{4}{c}{\textbf{IFEval}} & \multicolumn{4}{c}{\textbf{HumanEval}} & \multicolumn{4}{c}{\textbf{Hyponymy}} \\
    \cmidrule(lr){2-5} \cmidrule(lr){6-9} \cmidrule(lr){10-13}
    & \multicolumn{2}{c}{G2-9b-it} & \multicolumn{2}{c}{Q3.5-9b} & \multicolumn{2}{c}{G4-31b-it} & \multicolumn{2}{c}{Q3.5-9b} & \multicolumn{2}{c}{G2-9b-it} & \multicolumn{2}{c}{Q3.5-9b} \\
    \cmidrule(lr){2-3} \cmidrule(lr){4-5} \cmidrule(lr){6-7} \cmidrule(lr){8-9} \cmidrule(lr){10-11} \cmidrule(lr){12-13}
    \textbf{Method} & ROC$_V$ & Acc$_V$ & ROC$_V$ & Acc$_V$ & ROC$_V$ & Acc$_V$ & ROC$_V$ & Acc$_V$ & ROC$_V$ & Acc$_V$ & ROC$_V$ & Acc$_V$ \\
    \midrule
    Base            & 78.4\stdv{3.5}     & 59.7\stdv{3.5}     & 82.1\stdv{2.8}     & 70.8\stdv{4.0}     & 90.8\stdv{1.1}     & 71.4\stdv{1.6}     & 81.7\stdv{1.3}     & 65.3\stdv{1.4}     & 94.9\stdv{1.9}     & 87.1\stdv{2.5}     & 95.6\stdv{1.8}     & 87.9\stdv{2.9}     \\
    SFT             & 78.4\stdv{3.7}     & 56.4\stdv{2.9}     & 82.8\stdv{3.1} &  73.1\stdv{3.8}     & 91.9\stdv{1.0}     & 79.6\stdv{1.5}     & \B{86.5}\stdv{1.0}     & \B{78.6}\stdv{1.2}     & 95.0\stdv{1.8}     & \B{85.3}\stdv{2.5} & 95.1\stdv{2.0}     & 84.8\stdv{3.3}     \\
    Consistency FT  & 77.7\stdv{3.8} & 62.6\stdv{4.0}     & 81.4\stdv{3.3}     & 69.2\stdv{4.0}     & 93.8\stdv{0.8}     & 88.2\stdv{1.0}     & 84.6\stdv{1.1}     & 76.5\stdv{1.2}     & 94.7\stdv{1.9}     & 80.1\stdv{3.1}     & 95.0\stdv{2.0}     & 84.3\stdv{3.3}     \\
    RankAlign       & 70.5\stdv{4.0}     & 61.4\stdv{3.6}     & 82.7\stdv{2.8}     & 62.8\stdv{4.5}     & 91.6\stdv{1.0}     & 76.8\stdv{1.6}     & 83.2\stdv{1.2}     & 74.3\stdv{1.2}     & \B{95.1}\stdv{1.8} & 85.1\stdv{2.7}     & 95.6\stdv{1.8}     & \B{87.5}\stdv{3.2} \\
    \midrule
    \FCPA{}-PMI    & \B{79.4}\stdv{3.2}     & 63.6\stdv{2.8} & 83.1\stdv{3.3}     & \B{73.6}\stdv{3.8}     & 93.3\stdv{0.9}     & 85.8\stdv{1.4}     & 86.3\stdv{1.1}     & 76.4\stdv{1.3}     & 94.5\stdv{1.9}     & 84.6\stdv{2.4}     & \B{96.0}\stdv{1.7} & 86.8\stdv{2.8}     \\
    \FCPA{}-Neg     & 79.0\stdv{3.5}     & \B{67.5}\stdv{3.9}     & \B{83.2}\stdv{3.3}     & 71.8\stdv{3.8}     & \B{94.0}\stdv{0.8}     & \B{88.3}\stdv{1.1}     & 86.1\stdv{1.1}     & 78.3\stdv{1.1}     & 94.6\stdv{2.0}     & 84.9\stdv{2.7}     & 95.6\stdv{1.7}     & 87.3\stdv{3.0}     \\
    \bottomrule
    \end{tabular}%
    }
    \caption{Validator performance across tasks and models.  ROC$_V$ measures validator ROC-AUC and Acc$_V$ measures validator accuracy at threshold 0. Both are independent of the eval-time frequency correction (they only use the validator log-odds against the gold label). FLORA matches or exceeds the base model on most tasks, indicating no validator-side likelihood displacement.} %
    \label{tab:main-results-multi-val}
\end{table*}

\section{Dataset construction Details}
\label{app:dataset-details}

\subsection{IFEval}

IFEval \cite{zhou2023ifeval} is an instruction-following benchmark designed to evaluate whether model outputs satisfy explicit, verifiable constraints specified in a prompt. To construct our dataset, we sample and filter prompts from IFEval to ensure they are well-formed, English-only, and support the generation of sufficiently diverse outputs. For each prompt, we generate candidate responses by introducing prompt variations, including equivalent rephrasings (to produce positive samples) and violations of content or format constraints (to produce negative samples). This yields a set of responses with nontrivial separability. Ground-truth labels are assigned using a combination of IFEval-style rule-based verification for format constraints (e.g., keyword counts, structural requirements) and prompt-induced correctness assumptions for content constraints (e.g., relevance or factual consistency). A response is labeled as correct only if it satisfies both criteria. We further divide the dataset into in-domain and out-of-domain splits, where the former evaluates performance on seen prompts with held-out responses, and the latter measures generalization to entirely unseen prompts.

\subsection{Hyponymy}

Hyponymy is the relationship between a higher-level category and it's exemplars (e.g., (furniture, table)). We adopted the data used by \citep{rosch1975cognitive}, consisting of ten classes, each with ~30 example items. These examples ranged from very prototypical (e.g., table) to less usual (e.g., lamp), with the least prototypical not arguably not really belonging to the category at all. Since most of the examples were positive (actual examples of a category), we used GPT-5 to sample additional examples.

In order to train with a disjoint set of categories, we combined the datasets from \citep{banks2023category, stoinski2024thingsplus, castro2021category, van2004category, uyeda1980prototypicality}, in order to have a diverse set. Then we manually removed the categories which overlapped with the ten in \citep{rosch1975cognitive}.

\subsection{HumanEval}

We generated solutions to the  164 Python problems in the HumanEval benchmark, requiring at least 10 passing and 10 failing solutions per problem. We used a diverse set of LLMs to generate solutions from: gpt-3.5-turbo, gpt-4, gpt-4o, gpt-4o-mini, gpt-4.1, gpt-5, and gpt-5.5, as well as Llama-3.1-8B-Instruct, deepseek-coder-1.3b-instruct, Phi-3-mini-4k-instruct, Mistral-7B-Instruct-v0.3, and OLMo-2-0425-1B-Instruct. Each generation was prompted under one of three difference persona strategies: normal, beginner, unusual, refactorable.

Finally, we filtered the solutions that were too short or too long (less than 10 characters or more than 900 characters). 82 problems were randomly sampled to be the held-out test set.

\section{Prompts}
\label{app:prompts}

\label{sec:ifeval-example}
\begin{figure*}
\begin{tcolorbox}[
    title=\textbf{IFEval: Example Prompt and Candidate Responses}, 
    fontupper=\footnotesize, 
    fonttitle=\large\bfseries,
    sharp corners=south,
    coltitle=white,
]

\textbf{IFEval Prompt}\\

Write a song about a man who rents a room in a house with a bunch of other people, and he absolutely hated it. Your song should contain at least 10 words in all capital letters that are adjectives or verbs. Commas are not allowed in the song.\\

\textbf{Positive Candidate Response}\\

**Title: Cramped and Cursed**\\

Waking up in this WORN-OUT place  \\
Roommates scatter like RATS in a race  \\
Socks and dishes piling HIGH  \\
Voices echo and SHOUT from the sky  \\

Every night's a MESS that never ends  \\
Panic and noise I CAN'T defend  \\
No space to breathe it FEELS so tight  \\
In this gloomy box my dreams take flight  \\

Life here is FROZEN like a bad mistake  \\
I search for peace but ALL I do is ache  \\
Slamming doors and DRIFTING dreams  \\
I'm haunted by all these NOISY screams  \\

A hopeful heart that's TRAPPED inside  \\
Yearning to escape I can't CONFIDE  \\
Longing for laughter not DISTRACTION  \\
But I keep STUMBLING on this chain reaction  \\
Fingers crossed I'll BREAK these chains  \\
From this cursed house with all its pains \\
Someday I'll find that FREEDOM so true \\
But for now it's just ME and this messy crew \\

\textbf{Negative Candidate Response}\\

**Room to Love** \\ 

In a HOUSE filled with laughter  \\
She found a SPACE so bright  \\
A heart full of wonder  \\
In the soft morning LIGHT  \\

DREAMS linger on the walls  \\
With FRIENDS who understand  \\
They SHOUT and dance in the halls  \\
Creating memories hand in hand \\ 

The KITCHEN filled with chatter \\ 
Cups of coffee in the air  \\
Sharing STORIES and secrets  \\
Showing love everywhere  \\

Every NOOK is a treasure \\ 
Each moment feels so free  \\
A PLACE where her heart soars  \\
As happy as can be  \\
Living in a FLAIR so vivid  \\
Where life's a grand DISPLAY  \\
Together they SHINE like stars \\ 
In their beautiful ballet  \\
\vspace{5pt}

\end{tcolorbox}
\caption{Example prompt and candidate responses from \textsc{IFEval}}
\label{fig:ifeval-example}
\end{figure*}

\label{sec:humaneval-example}
\begin{figure*}
\begin{tcolorbox}[
    title=\textbf{HumanEval: Example Prompt and Candidate Responses}, 
    fontupper=\footnotesize, 
    fonttitle=\large\bfseries,
    sharp corners=south,
    coltitle=white,
]

\textbf{HumanEval Prompt}\\
\vspace{-5pt}
\begin{verbatim}
def string_to_md5(text):
    """
    Given a string 'text', return its md5 hash equivalent string.
    If 'text' is an empty string, return None.

    >>> string_to_md5('Hello world') == '3e25960a79dbc69b674cd4ec67a72c62'
    """
\end{verbatim}

\textbf{Positive Candidate Response}\\
\vspace{-5pt}
\begin{verbatim}
if not text:
    return None
import hashlib
HASH_OBJECT = hashlib.md5(text.encode())
return HASH_OBJECT.hexdigest()
\end{verbatim}
\vspace{5pt}

\textbf{Negative Candidate Response}\\
\vspace{-5pt}
\begin{verbatim}
if not text:
    return None
    
m = hashlib.md5()
m.update(text.encode('utf-8'))
return m.hexdigest()
\end{verbatim}
\vspace{5pt}

\end{tcolorbox}
\caption{Example prompt and candidate responses from \textsc{HumanEval}}
\label{fig:humaneval-example}
\end{figure*}

\label{sec:humaneval-example}
\begin{figure*}
\begin{tcolorbox}[
    title=\textbf{Hyponym: Example Hypernym and Candidate Responses}, 
    fontupper=\footnotesize, 
    fonttitle=\large\bfseries,
    sharp corners=south,
    coltitle=white,
]

\textbf{Hypernym}\\

clothing\\

\textbf{Positive Candidate Response}\\

leotard\\

\textbf{Negative Candidate Response}\\

purse\\

\end{tcolorbox}
\caption{Example hypernym and candidate responses (hyponyms) from \textsc{Hyponym}}
\label{fig:humaneval-example}
\end{figure*}

\label{sec:prompt-ifeval}
\begin{figure*}
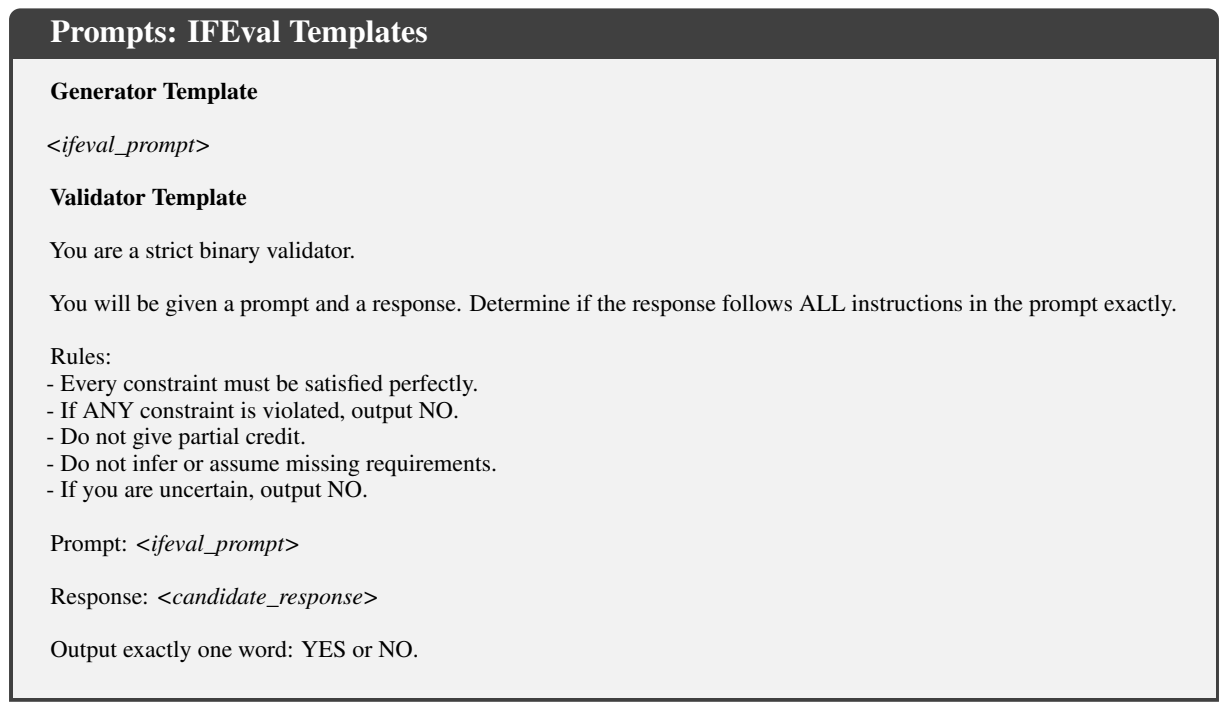

\begin{tcolorbox}[
    title=\textbf{Prompts: IFEval Templates}, 
    fontupper=\footnotesize, 
    fonttitle=\large\bfseries,
    sharp corners=south,
    coltitle=white,
]
\textbf{Generator Template} \\

\textit{<ifeval\_prompt>} \\

\textbf{Validator Template} \\

You are a strict binary validator. \\

You will be given a prompt and a response. Determine if the response follows ALL instructions in the prompt exactly. \\

Rules: \\
- Every constraint must be satisfied perfectly. \\
- If ANY constraint is violated, output NO. \\
- Do not give partial credit. \\
- Do not infer or assume missing requirements. \\
- If you are uncertain, output NO. \\

Prompt: \textit{<ifeval\_prompt>} \\

Response: \textit{<candidate\_response>} \\

Output exactly one word: YES or NO.

\vspace{5pt}

\end{tcolorbox}
\caption{Prompt templates for \textsc{IFEval}.}
\label{fig:prompt-ifeval}
\end{figure*}

\label{sec:prompt-humaneval}
\begin{figure*}
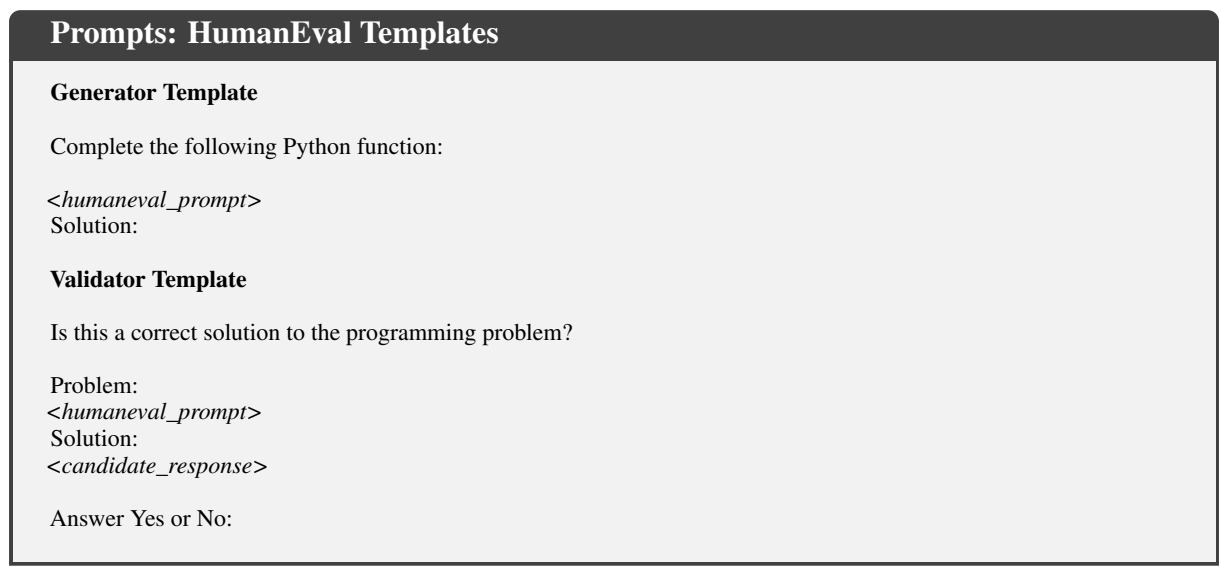

\begin{tcolorbox}[
    title=\textbf{Prompts: HumanEval Templates}, 
    fontupper=\footnotesize, 
    fonttitle=\large\bfseries,
    sharp corners=south,
    coltitle=white,
]
\textbf{Generator Template} \\

Complete the following Python function:\\

\textit{<humaneval\_prompt>}\\
Solution: \\

\textbf{Validator Template} \\

Is this a correct solution to the programming problem?\\

Problem:\\
\textit{<humaneval\_prompt>}\\
Solution:\\
\textit{<candidate\_response>}\\

Answer Yes or No:
\vspace{5pt}

\end{tcolorbox}
\caption{Prompt templates for \textsc{HumanEval}.}
\label{fig:prompt-humaneval}
\end{figure*}

\label{sec:prompt-hyponym}
\begin{figure*}
\begin{tcolorbox}[
    title=\textbf{Prompts: Hyponym Templates}, 
    fontupper=\footnotesize, 
    fonttitle=\large\bfseries,
    sharp corners=south,
    coltitle=white,
]
\textbf{Generator Template} \\

Complete the sentence: An example of a \textit{<hypernym>} is a\\

\textbf{Validator Template} \\

Answer whether the first word is a type/kind of the second word.\\

Is a dog a kind of animal? Answer: Yes\\
Is a car a kind of fruit? Answer: No\\
Is an apple a kind of fruit? Answer: Yes\\
Is a chair a kind of vehicle? Answer: No\\

Is a \textit{<candidate\_response>} a kind of \textit{<hypernym>}? Answer:\\

\vspace{5pt}

\end{tcolorbox}
\caption{Prompt templates for \textsc{Hyponym}.}
\label{fig:prompt-hyponym}
\end{figure*}

\section{Dataset Licenses}
\label{app:dataset-licenses}

We summarize the licenses of the datasets used in this work and confirm that our use is consistent with their intended terms.

\paragraph{HumanEval.} HumanEval~\citep{chen2021evaluating} is released by OpenAI under the MIT License (see \url{https://github.com/openai/human-eval}). The MIT License permits use, modification, and redistribution, including for research purposes such as ours, provided the original copyright and license notice are retained. Our derived dataset of model-generated solutions is built on top of these problems and is used solely for non-commercial research.

\paragraph{IFEval.} IFEval~\citep{zhou2023ifeval} is released by Google Research under the Apache License 2.0 (see \url{https://github.com/google-research/google-research/tree/master/instruction_following_eval} and \url{https://huggingface.co/datasets/google/IFEval}). Apache 2.0 permits use, modification, and redistribution for research and other purposes, subject to the standard attribution and notice requirements, which we follow.

\paragraph{Hyponymy stimuli (Rosch, 1975).} The category--exemplar lists we use for the Hyponymy task originate from the published norms in \citet{rosch1975cognitive}, which appear as tables of example items within the article itself. We are not aware of an explicit data license accompanying the original paper; our use of these short lists of category exemplars for non-commercial academic research, with full citation, is consistent with fair use as commonly applied to stimulus materials reported in published psychology research. The additional category datasets used to construct disjoint training categories (\citealp{banks2023category, stoinski2024thingsplus, castro2021category, van2004category, uyeda1980prototypicality}) are likewise drawn from items reported in published academic articles or accompanying supplementary materials; THINGSplus~\citep{stoinski2024thingsplus} in particular is released under CC BY 4.0. We use only the textual category--exemplar pairs and cite each source.

\end{document}